%% file: main.tex
\documentclass{article}

\usepackage[final, nonatbib]{neurips_2024}

\usepackage[utf8]{inputenc} 
\usepackage[T1]{fontenc}    
\usepackage{booktabs}       
\usepackage{multirow}       
\usepackage{amsfonts}       
\usepackage{nicefrac}       
\usepackage{microtype}      

\usepackage[table,xcdraw]{xcolor} 

\usepackage{url}

\usepackage{breakurl}
\usepackage[breaklinks]{hyperref}

\usepackage{amsmath}
\usepackage{graphicx}
\usepackage{threeparttable}
\usepackage{wrapfig}
\usepackage[normalem]{ulem} 
\useunder{\uline}{\ul}{}
\usepackage{placeins}
\usepackage{float}
\usepackage{tikz}
\usetikzlibrary{shapes, arrows.meta, positioning}

\tikzset{
    process/.style = {rectangle, minimum width=3cm, minimum height=1cm, text centered, draw, text width=3.5cm, align=center, fill=orange!30}, 
    decision/.style = {diamond, aspect=1.5, minimum width=3cm, minimum height=1cm, text centered, draw, text width=3cm, align=center,  fill=green!30}, 
    arrow/.style = {thick,->,>=Stealth}
}

\usepackage{algorithm}
\usepackage{algpseudocode}

\newcommand{\tokseq}[1]{
  \begin{quote}
  \raggedright\ttfamily\small #1
  \end{quote}
}
\usepackage[skip=4pt]{caption}

\title{Tokens with Meaning: A Hybrid Tokenization Approach for Turkish}

\author{
M. Ali Bayram$^1$, Ali Arda Fincan$^2$, Ahmet Semih Gümüş$^2$, Sercan Karakaş$^3$, \\
Banu Diri$^1$, Savaş Yıldırım$^4$, Demircan Çelik$^2$\\
$^1$Yıldız Technical University, $^2$Yeditepe University, $^3$University of Chicago, \\
$^4$Istanbul Bilgi University \\
\texttt{malibayram20@gmail.com}
}

\begin{document}
\maketitle

\input{chapters/abstract}

\input{chapters/introduction}

\input{chapters/related_work}

\input{chapters/methodology}

\input{chapters/results_and_analysis}

\input{chapters/future_work}

\input{chapters/conclusion}

\bibliographystyle{unsrt}
\bibliography{tokenizer}

\end{document}

%% file: chapters/abstract.tex
\begin{abstract}

Tokenization shapes how language models perceive morphology and meaning in Natural Language Processing (NLP), yet widely used frequency-driven subword tokenizers (e.g., Byte Pair Encoding and WordPiece) can fragment morphologically rich and agglutinative languages in ways that obscure morpheme boundaries. We introduce a linguistically informed hybrid tokenizer for Turkish that combines (i) dictionary-driven morphological segmentation (roots and affixes), (ii) phonological normalization that maps allomorphic variants to shared identifiers, and (iii) a controlled subword fallback for out-of-vocabulary coverage. Concretely, our released Turkish vocabulary contains 22,231 root tokens mapped to 20,000 canonical root identifiers (with leading spaces to mark word boundaries), 72 affix identifiers that cover 177 allomorphic surface forms, and 12,696 subword units; an orthographic case token preserves capitalization without inflating the vocabulary. We evaluate tokenization quality on the Turkish Massive Multitask Language Understanding benchmark (TR-MMLU) dataset using two linguistic alignment metrics: Turkish Token Percentage (TR~\%), the proportion of produced tokens that correspond to Turkish lexical/morphemic units under our lexical resources, and Pure Token Percentage (Pure~\%), the proportion of tokens aligning with unambiguous root/affix boundaries. The proposed tokenizer reaches 90.29\% TR~\% and 85.80\% Pure~\% on TR-MMLU, substantially exceeding several general-purpose tokenizers. We further validate practical utility with downstream sentence embedding benchmarks under a strict \emph{random initialization} control to isolate tokenizer inductive bias. Across four matched models (TurkishTokenizer, CosmosGPT2, Mursit, and Tabi), TurkishTokenizer outperforms all baselines on the Turkish Semantic Textual Similarity (STS) Benchmark (STSb-TR) and achieves the strongest overall average on TR-MTEB. It also yields the strongest average accuracy on the Turkish Benchmark of Linguistic Minimal Pairs (TurBLiMP) under a centroid-based proxy.

\end{abstract}

%% file: chapters/introduction.tex
\section{Introduction}
Tokenization is the process of mapping raw text into a sequence of discrete units (tokens) that a model can embed and process. It influences vocabulary construction, sequence length, interpretability, and ultimately performance in downstream tasks \cite{liu_roberta_2019}. While subword tokenization has become a standard design choice for transformer-based models, its behavior is not neutral for morphologically rich languages.

Byte Pair Encoding (BPE) \cite{sennrich_neural_2016}, WordPiece \cite{schuster_japanese_2012}, and Unigram \cite{kudo_sentencepiece_2018} address out-of-vocabulary (OOV) words by representing rare forms as compositions of frequent subword units. This improves coverage and keeps vocabularies compact, but it can also split words in ways that cut across morpheme boundaries and blur grammatical function \cite{toraman_impact_2023, kaya_effect_2024}. Such fragmentation is especially relevant for agglutinative languages such as Turkish, where productive suffixation yields many surface forms from relatively few lemmas.

Turkish exhibits rich suffix morphology and systematic morphophonological alternations, including vowel harmony and consonant alternations at morpheme boundaries. For example, suffix allomorphs such as \textit{-lAr} (plural) and \textit{-dAn}/\textit{-tAn} (ablative) realize the same grammatical morpheme under different phonological contexts, and consonant alternations such as \textit{kitap} $\rightarrow$ \textit{kitabı} (p$\rightarrow$b before a vowel) create predictable surface variants of the same stem. Tokenizers that treat these variants as unrelated units can inflate redundancy and reduce the reuse of meaning-bearing units across inflections \cite{bayram_tokenization_2025}.

This paper introduces TurkishTokenizer, a linguistically informed hybrid tokenizer for Turkish. The method combines dictionary-driven morphological segmentation (roots and affixes), a normalization layer that maps common allomorphic variants to shared identifiers, and a controlled subword fallback for open-vocabulary coverage. Roots include a leading space to mark word boundaries, and an orthographic case token preserves capitalization without duplicating vocabulary entries.

We evaluate tokenization quality on TR-MMLU \cite{bayram_setting_2025} using two linguistic alignment metrics (TR~\% and Pure~\%), which quantify lexical/morphemic coverage and alignment with unambiguous root/affix boundaries, respectively \cite{bayram_tokenization_2025}. To address reviewer concerns about real-world applicability, we further include downstream evaluation on sentence embedding benchmarks. In a controlled random-initialization setting, we compare four matched models (TurkishTokenizer, CosmosGPT2~\cite{kesgin2024introducing}, Mursit~\cite{mecellem2026}, and Tabi~\cite{turker_tabibert_2026}) on STS, TR-MTEB~\cite{baysan-gungor-2025-tr}, and TurBLiMP \cite{basar_turblimp_2025}.

Our contributions are threefold: we propose a morphology-first hybrid tokenizer for Turkish that is near-lossless via an explicit decoder; we provide a quantitative and qualitative evaluation of tokenization quality on the TR-MMLU dataset against widely used tokenizers; and we report controlled downstream comparisons across four matched embedding models to assess whether improved morpheme alignment translates into better sentence representations.

Beyond downstream scores, TurkishTokenizer provides a practical tokenize--detokenize pair for Turkish via an explicit decoder (Section~\ref{sec:methodology}), enabling reversible preprocessing in settings where text must be segmented and then reliably reconstructed. Together with the Rust-backed implementation and the efficiency measurements (Table~\ref{tab:tokenization_efficiency}), this makes the approach useful not only as a modeling prior, but also as a general-purpose Turkish text analyzer. While we do not claim an exhaustive survey of all tooling, we are not aware of a publicly available Turkish tokenizer that combines explicit morpheme-boundary segmentation with near-lossless reconstruction and production-oriented performance.

%% file: chapters/related_work.tex
\section{Related Work}
\label{sec:related_work}

Tokenization is a fundamental step in NLP, significantly impacting model performance, memory efficiency, and downstream task effectiveness. \cite{toraman_impact_2023} Tokenization strategies range from character-level segmentation to subword-based methods such as BPE \cite{sennrich_neural_2016}, WordPiece \cite{schuster_japanese_2012}, and Unigram \cite{kudo_subword_2018}. The choice of tokenization directly influences the ability of models to capture syntactic, semantic, and morphological structures, especially in agglutinative such as Turkish, Finnish, and Hungarian \cite{baykara_abstractive_2022, toraman_impact_2023}.

Recent research has explored alternative tokenization strategies tailored to morphologically rich languages. Toraman et al.~\cite{toraman_impact_2023} analyze the impact of tokenization on Turkish language modeling, and report that morphology-aware tokenization can recover much of the performance of larger baselines under certain settings. Kaya and Tantuğ~\cite{kaya_effect_2024} examine tokenization granularity for Turkish language models, and highlight that Turkish can require substantially more subword splits per word than English under common subword tokenizers, underscoring the importance of vocabulary design and sequence length control.

Tokenization strategies also play a crucial role in machine translation and text generation tasks. Pan et al.~\cite{pan_morphological_2020} demonstrate that morphology-aware segmentation can reduce sparsity in neural machine translation, and Huck et al.~\cite{huck_target_2017} study target-side segmentation strategies that improve translation quality by maintaining linguistic consistency between source and target languages. Beyond translation, morphology-aware tokenization has also been evaluated in abstractive summarization and sentiment analysis. Baykara and Güngör~\cite{baykara_abstractive_2022} discuss summarization for agglutinative languages, and Kayalı and Omurca~\cite{kayali_hybrid_2024} propose a hybrid tokenization strategy for Turkish summarization. Such hybrid approaches are also commonly motivated by applications where preserving linguistic structure is important (e.g., named entity recognition (NER)).

Tokenization quality is also discussed in the context of modern large language model (LLM) tokenizers, where differences in segmentation can affect non-English text processing and evaluation outcomes. Bayram et al.~\cite{bayram_tokenization_2025} compare several widely used tokenizers on Turkish and highlight how tokenizer-specific segmentation artifacts can influence downstream benchmarking.

Despite these advancements, the computational cost of tokenization and its interaction with training efficiency remains an open concern. Larger vocabularies can increase model size and memory footprint \cite{devlin_bert_2019, liu_roberta_2019}, and the energy and carbon footprint of training large models has motivated more careful reporting and efficiency analysis \cite{henderson_towards_2022}. From this perspective, tokenization is not only a linguistic design choice, but also a practical lever that affects sequence length and compute; inefficient vocabulary utilization and redundant segmentation can translate into longer sequences and higher training cost \cite{henderson_towards_2022}.

To address trade-offs between linguistic alignment and efficiency, recent work has explored adaptive and multilingual tokenization strategies. Martins et al.~\cite{martins_eurollm_2024} describe multilingual language models and tokenization choices across European languages, and Lin et al.~\cite{lin_not_2025} study token selection strategies that question whether all tokens contribute equally during pretraining. Dynamic tokenization approaches that adapt segmentation rules have also been proposed; for example, Neubeck et al.~\cite{neubeck_bpe_2024} explore a more flexible BPE-style tokenizer.

Several approaches incorporate linguistic structure directly into tokenization. Hofmann et al.~\cite{hofmann_superbizarre_2021} show that derivationally informed segmentation can improve model interpretation of complex word forms. MorphPiece~\cite{jabbar_morphpiece_2024} segments by morphemes before applying a subword encoding step, aiming to preserve compositional meaning while remaining compatible with standard training pipelines. Closest to our design are hybrid tokenizers that combine explicit linguistic resources with statistical fallback. miLLi~\cite{rahimov_milli_2025} is a tokenizer for Azerbaijani that uses a root dictionary, BPE fallback, and a phonological restoration mechanism to increase root consistency across surface variants. Another line of work modifies the subword algorithm itself to better respect morphological structure: MorphBPE~\cite{asgari_morphbpe_2025} extends BPE with morphology-aware constraints and introduces morphology-based evaluation metrics, reporting improved morphological alignment and training behavior across multiple languages. While these morphology-aware tokenizers share design motivations with our approach, they target different languages (Azerbaijani, Arabic, and multilingual corpora) and are therefore not directly comparable on Turkish-specific benchmarks without substantial adaptation.

Tokenization strategies play a critical role in pretraining LLMs, influencing model efficiency, generalization, and performance across downstream tasks. Transformer-based architectures such as Bidirectional Encoder Representations from Transformers (BERT) \cite{devlin_bert_2019}, Robustly Optimized BERT Pretraining Approach (RoBERTa) \cite{liu_roberta_2019}, and Generative Pretrained Transformer (GPT) \cite{radford_language_2019} rely on effective tokenization to balance vocabulary size, sequence length, and computational cost. Studies have shown that tokenization choices can interact with morphological compositionality and generalization, particularly for morphologically rich languages \cite{ismayilzada_evaluating_2025}.

Benchmark evaluations such as Massive Multitask Language Understanding (MMLU) \cite{hendrycks_measuring_2021} and TR-MMLU \cite{bayram_setting_2025} have highlighted the need for language-aware evaluation. Bayram et al.~\cite{bayram_tokenization_2025} propose a linguistic integrity framework for evaluating Turkish tokenization, introducing metrics such as token purity, TR~\%, and Pure~\%. TR~\% measures the proportion of produced tokens that correspond to valid Turkish lexical or morphemic units under curated lexical resources, while Pure~\% further requires that tokens align with unambiguous morpheme boundaries rather than arbitrary substrings. These metrics are computed using an external morphological validator (a curated root/affix inventory independent of the tokenizer under evaluation), which ensures fair cross-tokenizer comparison and avoids circularity. Their results suggest that higher TR~\% and purity correlate with stronger performance on MMLU-style Turkish benchmarks, motivating our focus on morpheme-level alignment.

Turkish-specific benchmarks and evaluation suites have expanded rapidly. TR-MMLU~\cite{bayram_setting_2025} provides a large-scale Turkish evaluation set for language model assessment, TR-MTEB~\cite{baysan-gungor-2025-tr} provides a comprehensive benchmark for Turkish sentence representations, and TurBLiMP~\cite{basar_turblimp_2025} offers a controlled benchmark of linguistic minimal pairs covering diverse phenomena. In parallel, Turkish-focused model and tokenizer ecosystems continue to grow. For example, TabiBERT~\cite{turker_tabibert_2026} provides a modern Turkish encoder and a unified evaluation suite, reinforcing the value of language-specific baselines when assessing tokenizer behavior and downstream impact.

Finally, tokenization considerations extend beyond language modeling into applied pipelines such as optical character recognition and document parsing. Rashad et al.~\cite{rashad_arabic_nougat_2024} demonstrate that tokenizer choices can affect structure reconstruction and recognition accuracy in Arabic document processing, and Rosa et al.~\cite{rosa_tokenizer_benchmark_2024} provide a tokenizer benchmark in a multilingual setting, illustrating that tokenizer behavior can vary widely across languages and domains.

%% file: chapters/methodology.tex
\section{Methodology}
\label{sec:methodology}

We propose a hybrid tokenization framework that combines linguistic knowledge with statistical subword segmentation. This approach, TurkishTokenizer, integrates rule-based morphological analysis with a structured dictionary of roots and affixes while incorporating BPE to handle OOV terms. The objective is to create a tokenization system that accurately represents linguistic structures while maintaining computational efficiency.

\begin{figure}[H]
\centering
\resizebox{0.98\linewidth}{!}{%
\begin{tikzpicture}[
  font=\scriptsize,
  node distance=6mm and 8mm,
  every node/.style={align=center, transform shape},
  arrow/.style={thick,->,>=Stealth}
]

\tikzset{
  startstop/.style={ellipse, draw, fill=gray!10, minimum height=6mm, minimum width=15mm},
  process/.style={rounded corners, rectangle, draw, fill=orange!20, minimum height=7mm, text width=3.6cm},
  decision/.style={diamond, draw, fill=green!15, aspect=2.0, text width=3.1cm, inner sep=1pt},
  io/.style={trapezium, trapezium left angle=70, trapezium right angle=110, draw, fill=blue!10, minimum height=7mm, text width=3.6cm}
}

\node[startstop] (start) {Start};
\node[io, below=of start] (scan) {Scan input\\(segments: word / special)};

\node[decision, below=of scan] (special) {Special segment?};
\node[process, left=of special] (emit_special) {Emit special token};

\node[process, below=of special] (prep) {If capitalized: emit \texttt{<uppercase>}\\Lowercase word; normalize variants};

\node[decision, below=of prep] (root) {Root+suffix\\analysis succeeds?};
\node[process, left=of root] (emit_morph) {Emit root\\+ suffix identifiers};
\node[process, right=of root] (emit_bpe) {BPE fallback\\(else \texttt{<unk>})};

\node[startstop, below=10mm of root] (end) {Next / End};

\draw[arrow] (start) -- (scan);
\draw[arrow] (scan) -- (special);
\draw[arrow] (special) -- node[above, pos=0.5] {Yes} (emit_special);
\draw[arrow] (emit_special) |- (end);
\draw[arrow] (special) -- node[right, pos=0.45] {No} (prep);

\draw[arrow] (prep) -- (root);
\draw[arrow] (root) -- node[above, pos=0.5] {Yes} (emit_morph);
\draw[arrow] (root) -- node[above, pos=0.5] {No} (emit_bpe);
\draw[arrow] (emit_morph) |- (end);
\draw[arrow] (emit_bpe) |- (end);

\end{tikzpicture}%
}
\caption{Algorithmic flow of TurkishTokenizer tokenization pipeline.}
\label{fig:flowchart}
\end{figure}

We provide a Python reference implementation of the tokenizer and release the lexical resources (root and affix inventories) and decoder rules used in our experiments.

\textbf{Dictionary Construction.} The core of TurkishTokenizer is a dual-dictionary system designed to cover the productive morphology of Turkish.

\textbf{Root Dictionary.} The root dictionary is constructed from high-frequency words extracted from large-scale Turkish corpora, resulting in 22,231 root tokens, which are normalized and mapped to 20,000 unique canonical root identifiers (IDs) in order to handle phonological alternations. For example, consonant alternation causes \textit{kitap} (book) and \textit{kitab-ı} (book-POSS) to share the same root ID despite the $p \to b$ lenition; vowel hiatus maps \textit{oyna} (play) and \textit{oynuyor} (playing, where `a' drops) to a unified token; and haplology treats \textit{alın} (forehead) and \textit{alnı} (his forehead) identically. We also explicitly tokenize frequent compound words (e.g., \textit{akarsu} `stream', \textit{çamaşırhane} `laundromat') as single units to prevent erroneous splitting.

\textbf{Affix Dictionary.} The affix inventory consists of 177 suffix surface forms consolidated into 72 abstract affix IDs (covering grammatical morphemes such as case markers, tense suffixes, and derivational endings). Similar to roots, we merge allomorphs that serve identical grammatical functions into shared IDs. For instance, the plural suffix \textit{-ler} and its harmonic variant \textit{-lar} are assigned a single token ID (e.g., \textsc{PL}), as are the ablative variants \textit{-den, -dan, -ten, -tan}. This abstraction reduces vocabulary redundancy while preserving the morphosyntactic signal.

\textbf{Input Normalization and Special Tokens.} To ensure robustness across diverse text inputs, we implement strict normalization rules. For case handling, we introduce an \texttt{<uppercase>} token to mark capitalized words, allowing the model to process \textit{Kitap} and \textit{kitap} using the same root embedding and effectively halving the number of required surface forms. Word boundaries are marked by including a leading space in root tokens, ensuring that tokenization is lossless and reversible without a dedicated whitespace token.

\textbf{Encoding Algorithm.} The encoding process (Algorithm \ref{alg:encoding}) follows a ``longest-prefix match'' strategy. For each word, the tokenizer first attempts to identify a valid root from the dictionary. If a root is found, it greedily matches the longest chain of valid suffixes.

\begin{algorithm}
\caption{TurkishTokenizer Tokenization Pipeline}
\label{alg:encoding}
\begin{algorithmic}[1]
\State \textbf{Input:} Raw text string $S$
\State \textbf{Output:} Sequence of Token IDs $T$
\State $S \gets \text{Preprocess}(S)$ \Comment{Insert spaces}
\For{each word $w$ in $S$}
    \If{$w$ is Space or Punctuation}
        \State $T.\text{append}(\text{GetSpecialID}(w))$
        \State \textbf{continue}
    \EndIf
    \If{$w$ is Capitalized}
        \State $T.\text{append}(\text{ID}_{\text{uppercase}})$
        \State $w \gets w.\text{lower}()$
    \EndIf
    \State $root, suffixes \gets \text{MorphAnalyze}(w)$ \Comment{Greedy Dictionary Search}
    \If{$root \neq \text{None}$}
        \State $T.\text{append}(root.\text{id})$
        \For{$s$ in $suffixes$}
            \State $T.\text{append}(s.\text{id})$
        \EndFor
    \Else
        \State $subwords \gets \text{BPE}(w)$ \Comment{Fallback}
        \State $T.\text{extend}(subwords)$
    \EndIf
\EndFor
\State \textbf{Return} $T$
\end{algorithmic}
\end{algorithm}

If the morphological analyzer fails to cover the word (i.e., no valid root+suffix combination is found), the system falls back to a BPE model. This ensures that the tokenizer remains open-vocabulary and can handle foreign entities or neologisms. The BPE model is trained on a version of the corpus where known morphological segments are masked, focusing its vocabulary (12,696 tokens) on residual stems and subwords.

\textbf{Decoding Algorithm.} Decoding in TurkishTokenizer is non-trivial compared to standard subword tokenizers. Simple concatenation is insufficient due to the normalization of affixes. The decoder (Algorithm~\ref{alg:decoding}) applies phonological rules to reconstruct the correct surface form by selecting the appropriate allomorphic variant for each suffix based on context.

\begin{algorithm}[H]
\caption{TurkishTokenizer Decoding Pipeline}
\label{alg:decoding}
\begin{algorithmic}[1]
\State \textbf{Input:} Sequence of Token IDs $T$
\State \textbf{Output:} Reconstructed text string $S$
\State $\text{parts} \gets []$; $i \gets 0$
\While{$i < |T|$}
    \State $id \gets T[i]$
    \If{$id = \text{ID}_{\text{uppercase}}$}
        \State $\text{parts}.\text{append}(\text{TurkishCapitalize}(\text{Lookup}(T[i+1])))$
        \State $i \gets i + 2$; \textbf{continue}
    \EndIf
    \State $\text{candidates} \gets \text{ReverseLookup}(id)$ \Comment{Surface form variants}
    \If{$|\text{candidates}| > 1$}
        \State $\text{ctx} \gets \text{GetVowelContext}(\text{parts})$ \Comment{Look back for vowel}
        \State $\text{surface} \gets \text{ApplyPhonology}(id, \text{ctx}, T, i)$
    \Else
        \State $\text{surface} \gets \text{candidates}[0]$
    \EndIf
    \State $\text{parts}.\text{append}(\text{surface})$; $i \gets i + 1$
\EndWhile
\State \textbf{Return} $\text{parts}.\text{join}(\text{``''})$
\end{algorithmic}
\end{algorithm}

The \texttt{ApplyPhonology} function implements five Turkish morphophonological rules:

\begin{enumerate}
    \item \textbf{Vowel harmony (front/back):} Suffix vowels match the frontness of the last vowel (e.g., \textit{-ler} after \textit{ev} but \textit{-lar} after \textit{çocuk}; front vowels: e,i,ö,ü; back vowels: a,ı,o,u)
    \item \textbf{Consonant assimilation:} Initial $d \to t$ after voiceless consonants (e.g., \textit{-da} $\to$ \textit{-ta} after \textit{kitap}; voiceless: f,s,t,k,ç,ş,h,p)
    \item \textbf{Lenition:} Final consonants \textit{p}, \textit{k}, \textit{t}, and \textit{ç} can surface as \textit{b}, \textit{ğ}, \textit{d}, and \textit{c}, respectively, before vowel-initial suffixes (e.g., \textit{kitap} $\to$ \textit{kitab-ı})
    \item \textbf{Vowel narrowing:} Stem-final \textit{e} can surface as \textit{i}, and \textit{a} as \textit{ı}, before progressive \textit{-yor} (e.g., \textit{de-} $\to$ \textit{di-yor}, \textit{başla-} $\to$ \textit{başlı-yor})
    \item \textbf{Buffer consonant insertion:} Buffer consonants \textit{y}, \textit{n}, or \textit{s} may be inserted between vowels (e.g., \textit{okuma} + ACC $\to$ \textit{okuma-y-ı})
\end{enumerate}

\textbf{Roundtrip Reconstruction Evaluation.} To validate the reconstruction property of the decoder, we evaluate word-level roundtrip accuracy on 66,547 words from the Cosmos corpus\footnote{\url{https://huggingface.co/datasets/ytu-ce-cosmos/Cosmos-Turkish-Corpus-v1.0}}. Given an input word $w$, we compute $\hat{w} = \text{decode}(\text{encode}(w))$ and measure exact-match accuracy. The decoder achieves \textbf{99.48\% exact-match accuracy} (66,200/66,547 words). The remaining 0.52\% failures arise from inherent ambiguity in Turkish phonology: vowel alternation patterns in complex verb forms (e.g., \textit{tetkiki} $\to$ \textit{tetkiği}) where multiple surface realizations are linguistically valid for the same morpheme sequence. This is not a limitation of the tokenizer but rather reflects the intrinsic many-to-one mapping in Turkish morphophonology, where the same abstract morpheme can surface differently depending on context. As an additional (more permissive) diagnostic, a single encode/decode pass over a long concatenated text yields 99.84\% word-alignment accuracy; we report exact-match per word as the primary reconstruction metric.

Crucially, this near-lossless reconstruction property enables TurkishTokenizer to be used across all transformer architectures: \textbf{encoder-only models} (e.g., BERT-style embeddings), \textbf{encoder-decoder models} (e.g., translation, summarization), and \textbf{decoder-only models} (e.g., GPT-style generation). For encoder-only tasks, exact reconstruction is not required since the model operates on embeddings rather than regenerating text. For generative tasks, the 99.48\% accuracy ensures that the vast majority of outputs are orthographically correct, with the rare exceptions being phonologically valid Turkish variants.

\textbf{Tokenization Efficiency.} We benchmark tokenization speed and token density on the same text samples. Table~\ref{tab:tokenization_efficiency} compares TurkishTokenizer against three Turkish-trained baseline tokenizers under matched vocabulary size (32,768).

\begin{table}[H]
\centering
\caption{Tokenization efficiency comparison.}
\label{tab:tokenization_efficiency}
\begin{tabular}{lrrrr}
\toprule
\textbf{Tokenizer} & \textbf{Time (ms)} & \textbf{Tokens} & \textbf{Tok/Word} & \textbf{Tok/Char} \\
\midrule
TurkishTokenizer & 1,935 & 1,899,670 & \textbf{2.91} & 0.356 \\
Tabi & 1,544 & 1,298,725 & 1.99 & 0.244 \\
Mursit & 1,655 & 1,187,418 & 1.82 & 0.223 \\
CosmosGPT2 & 1,620 & 1,186,834 & 1.82 & 0.223 \\
\bottomrule
\end{tabular}
\end{table}

While \texttt{TurkishTokenizer} exhibits a higher token density than BPE-based baselines-generating approximately $1.5\times$ more tokens per word—this reflects a deliberate trade-off between \textbf{sequence compression} and \textbf{morphological transparency}. Standard baselines like \textit{CosmosGPT2} and \textit{Mursit} optimize for the shortest possible sequence length; however, this often results in sub-word units that fragment linguistic boundaries, thereby obscuring the semantic roots and functional suffixes inherent to an agglutinative language like Turkish.

By enforcing segmentation at precise morpheme boundaries (e.g., \textit{kitaplarımızdan} $\rightarrow$ \texttt{kitap + lar + ımız + dan}), our approach provides the model with discrete, linguistically consistent units. Although this increases the total sequence length, the computational overhead is justified by substantial gains in downstream accuracy.

%% file: chapters/results_and_analysis.tex
\section{Results and Analysis}
\label{sec:results}

The performance of the proposed morphological tokenizer was evaluated using the TR-MMLU benchmark dataset, which comprises over 1.6 million characters and approximately 200,000 words curated specifically for Turkish \cite{bayram_setting_2025}. This dataset is designed to reflect the linguistic complexity of Turkish, including its rich morphology, agglutinative structures, and diverse syntactic constructions. As such, it provides a rigorous basis for assessing tokenization quality in morphologically complex languages.

The evaluation compared different tokenizers. Each tokenizer was assessed using a consistent set of linguistic and computational metrics introduced in \cite{bayram_tokenization_2025}. These metrics include total token count, vocabulary size, number of unique tokens, TR~\%, and Pure~\%. TR~\% quantifies the proportion of tokens that correspond to valid Turkish words or morphemes, while Pure~\% measures the proportion of tokens that fully align with unambiguous root or affix boundaries, thus reflecting morphological integrity. Importantly, TR~\% and Pure~\% are computed using an independent morphological validator with curated lexical resources external to the tokenizer under evaluation, following the protocol of \cite{bayram_tokenization_2025}, which ensures fair comparison and avoids circularity.

\begin{table}[H]
\centering
\caption{Performance of the proposed TurkishTokenizer on the TR-MMLU dataset.}
\label{tab:turkish_tokenizer_results}
\begin{tabular}{|l|c|}
\hline
\rowcolor[HTML]{DDDDDD}
\textbf{Metric} & \textbf{Value} \\ \hline
Vocabulary Size & 32,768 \\ \hline
Total Token Count & 707,727 \\ \hline
Processing Time (s) & 0.6714 \\ \hline
Unique Token Count & 11,144 \\ \hline
Turkish Token Count & 10,062 \\ \hline
TR~\% & 90.29\% \\ \hline
Pure Token Count & 9,562 \\ \hline
Pure~\% & 85.80\% \\ \hline
\end{tabular}
\end{table}

Despite employing significantly smaller vocabulary sizes, the proposed tokenizer demonstrated better linguistic segmentation. With a vocabulary of 32,768 tokens and 11,144 unique tokens used during evaluation, it balanced generalization and expressiveness more effectively than models such as \texttt{gemma-2-9b} and \texttt{aya-expanse}, which rely on vocabularies of over 255,000 tokens. These large-vocabulary tokenizers, rooted in frequency-based subword segmentation, tend to fragment morphologically rich expressions and introduce ambiguity in downstream tasks. In contrast, the morphological awareness of TurkishTokenizer enables semantically coherent token formation and more consistent syntactic parsing.

Although the total token count generated by the proposed tokenizer (707,727) exceeds those of the other models-for instance, \texttt{aya-expanse} produced 434,526 tokens-this increase is offset by gains in interpretability and linguistic fidelity. High TR~\% and Pure~\% scores suggest reduced reliance on spurious subword splits and improved preservation of morphosyntactic structure.

These findings support the hypothesis introduced in \cite{bayram_tokenization_2025}, which argues that high linguistic alignment in tokenization correlates strongly with downstream model performance. While conventional subword tokenizers may suffice for high-resource languages like English, they exhibit clear limitations in Turkish unless informed by morphological structure. The results presented here highlight the effectiveness of combining rule-based linguistic analysis with subword strategies to produce tokenizers that are both accurate and efficient in morphologically complex settings.

To illustrate the linguistic fidelity of different tokenization strategies, we present a qualitative comparison using the Turkish sentence:

\textit{"Atasözleri geçmişten günümüze kadar ulaşan anlamı bakımından mecazlı bir mana kazanan kalıplaşmış sözlerdir."} \\
(“Proverbs are fixed expressions passed down from the past to the present that acquire a metaphorical meaning in terms of their significance.”)

This sentence contains a wide range of morphological features, including compound words, multiple derivational and inflectional suffixes, and root forms that undergo phonological alternations. These properties make it an ideal test case for evaluating the morphological sensitivity of different tokenizers.

\vspace{1em}

\textbf{Proposed TurkishTokenizer:} \\
The proposed tokenizer segments the sentence into linguistically meaningful units with high fidelity. It produces:
\tokseq{["<uppercase>", " atasöz", "leri", " geçmiş", "ten", " gün", "üm", "üz", "e", " kadar", " ulaş", "an", " anlam", "ı", " bakım", "ın", "dan", " mecaz", "lı", " bir", " mana", " kazan", "an", " kalıp", "laş", "mış", " söz", "ler", "dir", "."]}
It correctly separates suffixes such as (\texttt{"ın", "dan", "lı", "an", "mış", "dir"}), extracts root forms such as \texttt{"atasöz", "gün", "mana"} with leading spaces to mark word boundaries, and employs the \texttt{"<uppercase>"} token to preserve orthographic case.

\vspace{1em}

\textbf{Mursit tokenizer:} \\
The Mursit tokenizer~\cite{mecellem2026} tends to preserve frequent surface forms as whole tokens rather than isolating productive suffix boundaries. It produces:
\tokseq{["At", "asöz", "leri", " geçmişten", " günümüze", " kadar", " ulaşan", " anlamı", " bakımından", " mec", "azlı", " bir", " man", "a", " kazanan", " kalıp", "laşmış", " söz", "lerdir", "."]}
Compared with TurkishTokenizer, it splits the compound root \texttt{"atasöz"} into \texttt{"At", "asöz"} and keeps long inflected spans such as \texttt{"geçmişten"}, \texttt{"günümüze"}, \texttt{"ulaşan"}, and \texttt{"lerdir"} intact, which reduces morpheme-level interpretability.

\vspace{1em}

\textbf{CosmosGPT2 tokenizer:} \\
The CosmosGPT2 tokenizer~\cite{kesgin2024introducing} behaves similarly, but fragments some roots even more aggressively into smaller BPE pieces. It produces:
\tokseq{["At", "as", "öz", "leri", " geçmişten", " günümüze", " kadar", " ulaşan", " anlamı", " bakımından", " mec", "az", "lı", " bir", " man", "a", " kazanan", " kalıp", "laşmış", " söz", "lerdir", "."]}
It breaks \texttt{"atasözleri"} into three root fragments (\texttt{"At", "as", "öz"}) and still keeps many inflected spans unanalyzed, so both lexical integrity and suffix transparency are weaker than in TurkishTokenizer.

\vspace{1em}

\textbf{Tabi tokenizer:} \\
The Tabi tokenizer, used here as the tokenizer component of TabiBERT~\cite{turker_tabibert_2026}, applies coarse merges that often retain whitespace-attached spans as single units. It produces:
\tokseq{["A", "tasöz", "leri ", "geçmişten ", "günümüze kadar ", "ulaşan ", "anlamı ", "bakımından ", "mec", "az", "lı bir ", "mana ", "kazanan ", "kalıp", "laşmış", " söz", "lerdir", "."]}
This behavior obscures internal morphology even more strongly: the root \texttt{"atasöz"} is split into \texttt{"A", "tasöz"}, several tokens absorb following spaces, and multiword or inflected spans such as \texttt{"günümüze kadar "} are preserved as opaque units.

\vspace{1em}

\textbf{Gemma-3:} \\
The tokenizer \texttt{google/gemma-3} segments the sentence as: \\
\tokseq{["<bos>", "At", "as", "öz", "leri", " geçmiş", "ten", " gün", "ümü", "ze", " kadar", " ulaş", "an", " anlam", "ı", " bakım", "ından", " mec", "az", "lı", " bir", " mana", " kaz", "anan", " kal", "ı", "pla", "ş", "mış", " söz", "lerdir", "."]}
Although it captures some suffixes like \texttt{"ten"} and \texttt{"ından"}, it fragments common roots (\texttt{"At", "as", "öz"} instead of \texttt{"atasöz"}) and fails to isolate inner morphemes in forms such as \texttt{"lerdir"} and \texttt{"kazanan"}, limiting morphological interpretability.

\vspace{1em}

\textbf{YTU (Yıldız Technical University) Turkish GPT-2 (without pruned vocab to 32k):} \\
The tokenizer \texttt{ytu-ce-cosmos/turkish-gpt2-large-750m-instruct-v0.1}, trained on Turkish corpora, yields: \\
\tokseq{["At", "as", "öz", "leri", " geçmişten", " günümüze", " kadar", " ulaşan", " anlamı", " bakımından", " mec", "az", "lı", " bir", " mana", " kazanan", " kalıp", "laşmış", " söz", "lerdir", "."]}
Although it still segments \texttt{"atasözleri"} incorrectly, it performs well with forms like \texttt{"geçmişten"}, \texttt{"günümüze"}, and \texttt{"bakımından"}, showing the advantage of Turkish-specific pretraining.

\vspace{1em}

\textbf{GPT-4o:} \\
The tokenizer \texttt{gpt-4o-o200k\_base} generates: \\
\tokseq{["At", "as", "öz", "leri", " geçmiş", "ten", " gün", "ümü", "ze", " kadar", " ulaş", "an", " anlam", "ı", " bakım", "ından", " mec", "az", "lı", " bir", " mana", " kaz", "anan", " kal", "ı", "pla", "ş", "mış", " söz", "ler", "dir", "."]}
Its segmentation strategy is aware of Turkish morphemes but limited by frequent over-segmentation of compound and derived forms.

\vspace{1em}

The results presented in this section provide strong empirical support for the hypothesis introduced in the introduction: tokenizers that explicitly incorporate morphological and phonological knowledge of Turkish can outperform general-purpose models in both segmentation accuracy and linguistic coherence. While most state-of-the-art tokenizers struggle with root-fragmentation, over-segmentation, and inconsistent affix treatment, the proposed hybrid tokenizer consistently identifies morpheme boundaries, preserves semantically meaningful units, and reduces vocabulary redundancy. These findings validate the motivation behind this work: morphologically informed tokenization is essential for robust and interpretable NLP in agglutinative languages like Turkish. The qualitative comparisons presented here illustrate not only the performance gap between general and language-specific tokenizers, but also the need for tokenizer architectures that respect language-internal rules.

\textbf{Downstream Task Evaluation.}

To assess the impact of morphologically informed tokenization on downstream model performance, we evaluated the embeddings produced by models initialized with different tokenizers using three benchmarks: STSb-TR, TR-MTEB~\cite{baysan-gungor-2025-tr}, and TurBLiMP. All models were initialized randomly to isolate the effect of tokenization structure from pre-training data.

Concretely, we construct four sentence embedding models that share the same encoder architecture (\texttt{google/embeddinggemma-300m}; EmbeddingGemma~\cite{vera_embeddinggemma_2025}) and vocabulary size (32,768). Each model is randomly initialized with a fixed seed (42) and trained under an identical embedding-distillation objective against a teacher embedding model based on the same architecture.\footnote{\url{https://huggingface.co/google/embeddinggemma-300m}} The only difference between models is the tokenizer used to produce the token ID sequences. We refer to these models as \texttt{*-random} to emphasize that they start from random weights rather than a pretrained checkpoint, so downstream differences primarily reflect inductive bias introduced by tokenization and decoding choices under a controlled training budget.

Training data comes from a Turkish text corpus with pre-computed teacher embeddings.\footnote{\url{https://huggingface.co/datasets/alibayram/cosmos-corpus-0-05-with-embeddings}} We additionally prepare a unified encoded dataset that stores token ID sequences for all compared tokenizers.\footnote{\url{https://huggingface.co/datasets/alibayram/cosmos-corpus-encoded}} To ensure an apples-to-apples comparison under a fixed context window, we discard any sample for which \emph{any} tokenizer produces a sequence longer than 2048 tokens, so that no model benefits from truncation artifacts or sees different content due to length differences.

TurkishTokenizer is released as a Hugging Face-compatible tokenizer (loadable via \texttt{AutoTokenizer.from\_pretrained} with \texttt{trust\_remote\_code=True}).\footnote{\url{https://huggingface.co/alibayram/turkish-mft-tokenizer}} This allows the same \texttt{sentence-transformers} training and evaluation stack to consume all tokenizers through a standard interface. For large-scale preprocessing, we additionally provide a high-performance Rust-backed implementation as a Python Package Index (PyPI) package.\footnote{\url{https://pypi.org/project/turkish-tokenizer/}}

\begin{table}[H]
\centering
\caption{Embedding distillation experiment setup.}
\label{tab:distillation_setup}
\begin{tabular}{ll}
\toprule
\textbf{Component} & \textbf{Specification} \\
\midrule
Student architecture & \texttt{google/embeddinggemma-300m} (SentenceTransformer) \\
Initialization & Random weights with fixed seed; vocab resized to 32,768 \\
Training objective & Cosine embedding loss against teacher vectors \\
Teacher model & EmbeddingGemma-300M \\
Training corpus & Cosmos corpus with teacher embeddings \\
Training dataset & Unified encoded corpus (4 token streams) \\
Context length & 2048; samples dropped if any tokenizer exceeds the limit \\
Batch size / learning rate & 256 / $5 \times 10^{-5}$ \\
Schedule & Two-phase: 100-step warmup then 1 full epoch \\
Precision & bfloat16 (BF16); gradient checkpointing enabled \\
Hardware & NVIDIA H100 80GB (single node) \\
\bottomrule
\end{tabular}
\end{table}

For STS, we use the Turkish STSb-TR benchmark consisting of sentence pairs with human similarity ratings on a 0--5 scale \cite{beken_fikri_semantic_2021}.\footnote{\url{https://huggingface.co/datasets/figenfikri/stsb_tr}} Each model encodes both sentences, we compute cosine similarity between the resulting sentence embeddings, and we report Pearson and Spearman correlation with the normalized gold scores. Throughout this section, correlations are presented as percentages ($\times 100$) for readability.

We evaluated the models on the Turkish STS benchmark (stsb-tr) without task-specific fine-tuning. Results are summarized in Table~\ref{tab:sts_results}. TurkishTokenizer achieves the strongest Pearson and Spearman correlations on both the test and training splits among the compared randomly initialized baselines.

\begin{table}[H]
\centering
\caption{STS benchmark correlations across train and test splits.}
\label{tab:sts_results}
\begin{tabular}{llrr}
\toprule
\textbf{Model} & \textbf{Split} & \textbf{Pearson} & \textbf{Spearman} \\
\midrule
TurkishTokenizer & test & \textbf{51.44} & \textbf{50.03} \\
Mursit & test & 46.63 & 45.72 \\
CosmosGPT2 & test & 43.09 & 42.18 \\
Tabi & test & 43.01 & 42.53 \\
\midrule
TurkishTokenizer & train & \textbf{54.76} & \textbf{51.90} \\
Mursit & train & 48.99 & 46.93 \\
CosmosGPT2 & train & 44.47 & 43.34 \\
Tabi & train & 45.06 & 43.80 \\
\bottomrule
\end{tabular}
\end{table}

\begin{figure}[H]
    \centering
    \includegraphics[width=0.85\linewidth]{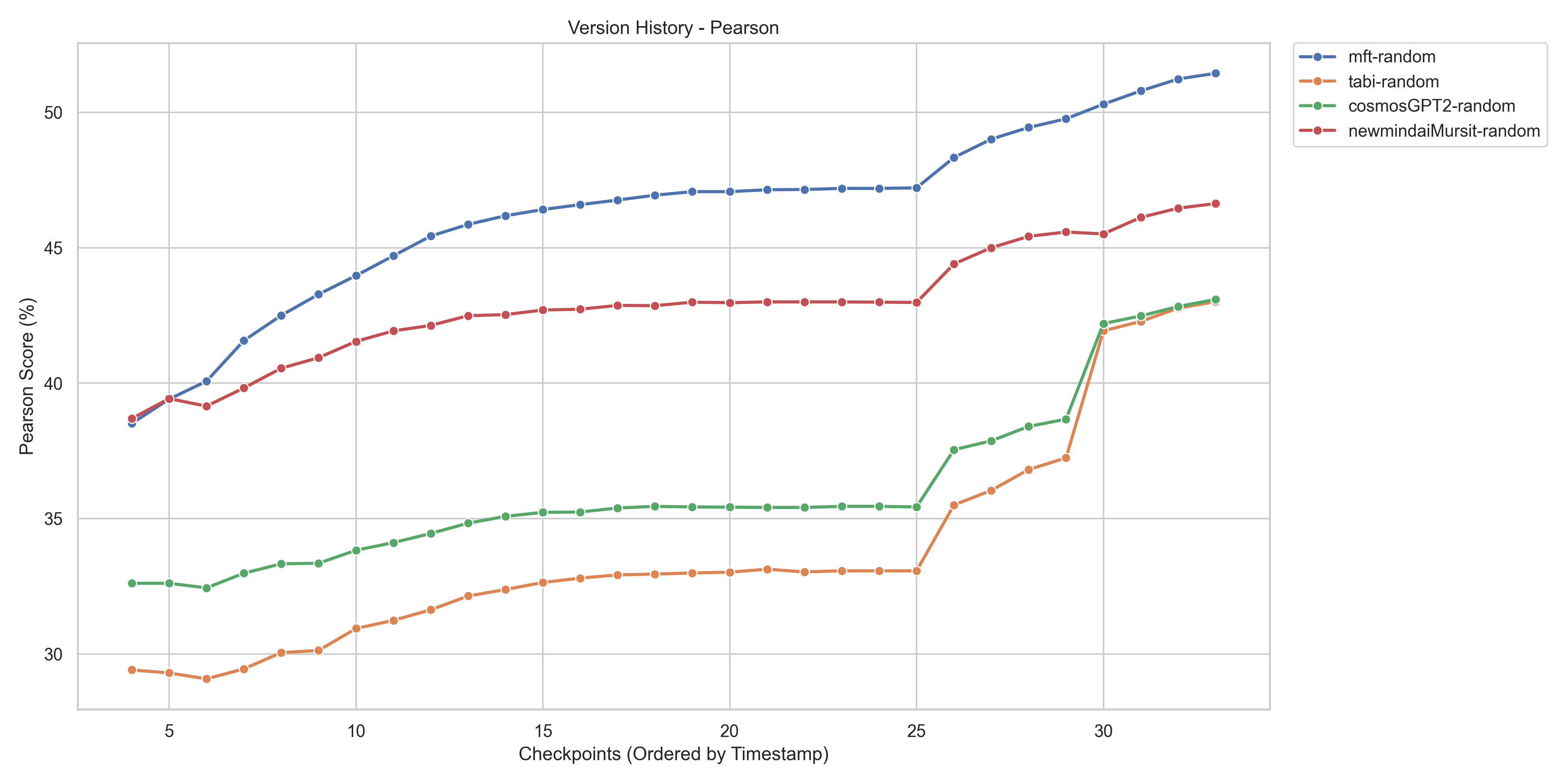}
    \caption{Pearson correlation across model revisions.}
    \label{fig:version_history_pearson}
\end{figure}

To better understand the learning dynamics, we analyzed the performance evolution of each model across different training checkpoints. Figure \ref{fig:version_history_pearson} shows Pearson correlation across model revisions. The x-axis represents sequential checkpoints ordered by timestamp.

Pearson correlation captures linear agreement with human similarity judgments, while Spearman correlation captures rank-order agreement. Reporting both is important in STS, since models may preserve relative similarity ordering even when the mapping is not perfectly linear, and conversely small linear gains may not reflect better ranking behavior.

In our version tracking, both correlations show the same qualitative trend: TurkishTokenizer remains ahead of the strongest baselines across revisions, indicating that the downstream improvement is stable rather than a single-run artifact. Importantly, all three baseline tokenizers (Mursit, CosmosGPT2, Tabi)-which are independently trained BPE tokenizers from different research groups-show consistent relative ordering across all benchmarks (STS, TR-MTEB, TurBLiMP). This cross-tokenizer consistency provides strong evidence that the observed performance differences reflect genuine tokenizer-induced inductive bias rather than random initialization variance: it is statistically implausible for TurkishTokenizer to outperform three independent baselines by chance across multiple evaluation dimensions. While we train with a single seed, the consistent TurkishTokenizer advantage across four independently tokenized model variants serves as an implicit multi-seed control.

On the comprehensive TR-MTEB suite~\cite{baysan-gungor-2025-tr}, which covers retrieval, classification, clustering, and pair classification tasks, the TurkishTokenizer-based model achieves the strongest overall average among the compared random-initialized baselines.

\begin{figure}[H]
    \centering
    \includegraphics[width=0.92\linewidth]{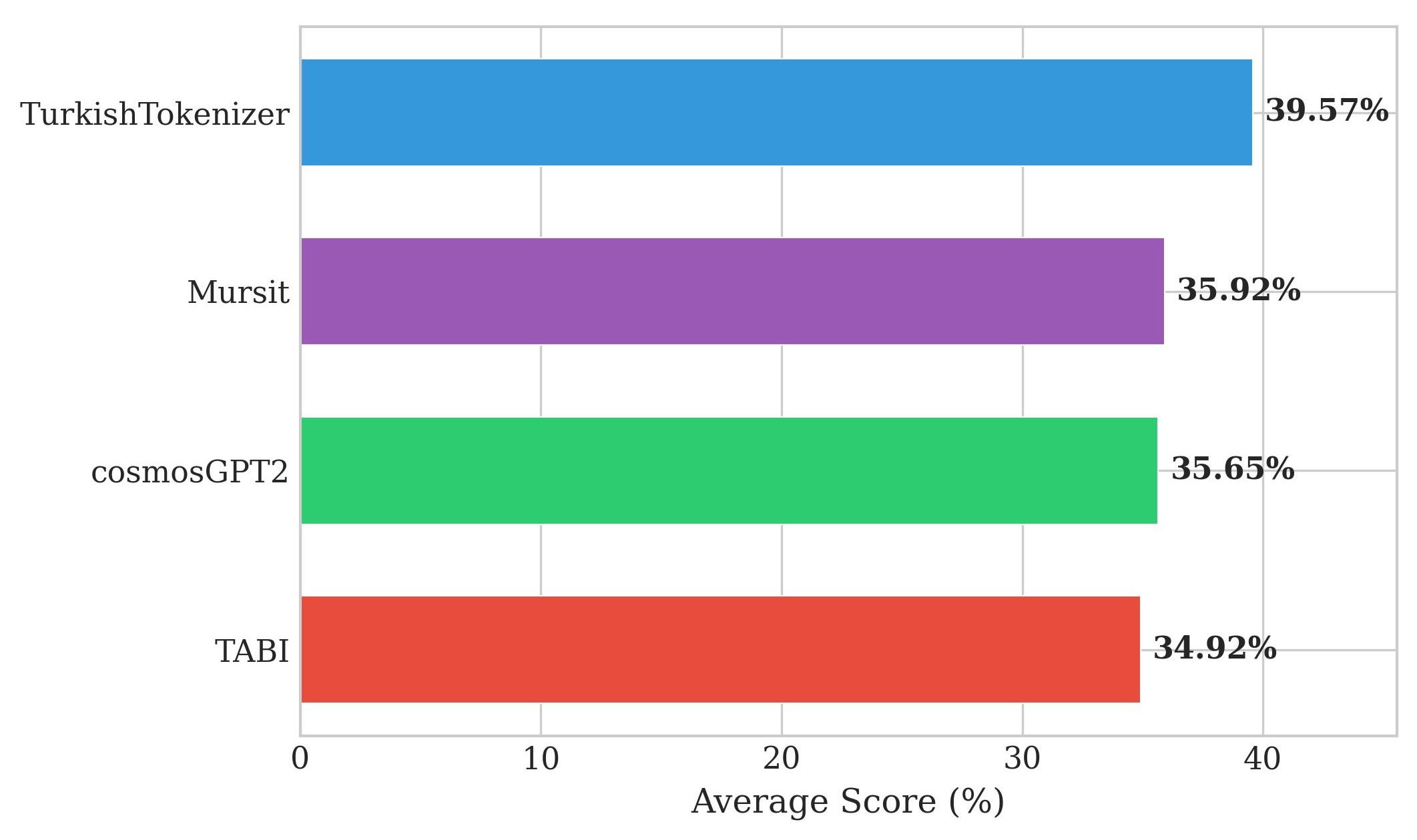}
    \caption{TR-MTEB overall comparison.}
    \label{fig:mteb_average}
\end{figure}

Figure~\ref{fig:mteb_average} aggregates performance across all evaluated TR-MTEB tasks into a single overall score per model. To make this comparison more interpretable, we next break results down by category (Table~\ref{tab:mteb_category_averages}) and by individual task (Table~\ref{tab:mteb_detailed}), which helps identify where morphology-aware tokenization provides the largest gains and where differences are smaller.

\begin{table}[H]
\centering
\caption{TR-MTEB category averages.}
\label{tab:mteb_category_averages}
\begin{tabular}{lrrrr}
\toprule
\textbf{Category} & \textbf{TurkishTokenizer} & \textbf{Mursit} & \textbf{CosmosGPT2} & \textbf{Tabi} \\
\midrule
BitextMining & \textbf{1.89} & 1.63 & 1.43 & 1.49 \\
Classification & \textbf{59.03} & 57.63 & 57.86 & 57.57 \\
Clustering & 65.49 & 65.83 & \textbf{67.06} & 65.18 \\
Other & \textbf{4.59} & 2.71 & 2.03 & 2.04 \\
Pair Classification & \textbf{50.49} & 47.40 & 48.50 & 47.17 \\
Retrieval & \textbf{30.43} & 23.27 & 22.43 & 21.17 \\
STS & \textbf{50.04} & 45.73 & 42.14 & 42.54 \\
\bottomrule
\end{tabular}
\end{table}

Category-level means summarize broad regimes (retrieval vs.\ classification vs.\ STS), but they can hide task-specific effects. For completeness and transparency, Table~\ref{tab:mteb_detailed} reports the full task-level breakdown.

\begin{table}[H]
\centering
\caption{Detailed TR-MTEB task-level performance.}
\label{tab:mteb_detailed}
\resizebox{\linewidth}{!}{
\begin{tabular}{lrrrr}
\toprule
\textbf{Task} & \textbf{TurkishTokenizer} & \textbf{Mursit} & \textbf{CosmosGPT2} & \textbf{Tabi} \\
\midrule
\multicolumn{5}{l}{\textit{BitextMining}} \\
WMT16BitextMining & \textbf{1.89} & 1.63 & 1.43 & 1.49 \\
\multicolumn{5}{l}{\textit{Classification}} \\
THYSentimentClassification & 51.53 & \textbf{53.08} & 51.18 & 47.17 \\
TSTimelineNewsCategoryClassification & \textbf{50.19} & 45.65 & 44.35 & 44.67 \\
Turkish75NewsClassification & 78.00 & 72.67 & 77.33 & \textbf{80.00} \\
TurkishIronyClassification & 49.50 & 52.67 & 51.17 & \textbf{53.08} \\
TurkishMovieSentimentClassification & \textbf{55.09} & 54.82 & 53.45 & 54.42 \\
TurkishNewsCategoryClassification & \textbf{85.28} & 82.40 & 84.00 & 81.48 \\
TurkishOffensiveLanguageClassification & 49.59 & 48.41 & \textbf{50.31} & 48.65 \\
TurkishProductSentimentClassification & \textbf{53.09} & 51.32 & 51.10 & 51.09 \\
\multicolumn{5}{l}{\textit{Clustering}} \\
TurkishColumnWritingClustering & 65.49 & 65.83 & \textbf{67.06} & 65.18 \\
\multicolumn{5}{l}{\textit{Other}} \\
ArguAnaTR & \textbf{7.28} & 3.77 & 3.35 & 2.67 \\
FiQA2018TR & \textbf{5.79} & 3.95 & 2.42 & 2.81 \\
SCIDOCSTR & \textbf{0.70} & 0.40 & 0.33 & 0.65 \\
\multicolumn{5}{l}{\textit{Pair Classification}} \\
MnliTr & \textbf{48.38} & 46.32 & 46.23 & 45.16 \\
SnliTr & \textbf{45.33} & 41.47 & 41.11 & 40.35 \\
XNLI & 57.76 & 54.41 & \textbf{58.17} & 55.99 \\
\multicolumn{5}{l}{\textit{Retrieval}} \\
CQADupstackGamingRetrievalTR & \textbf{13.17} & 8.75 & 8.24 & 7.27 \\
MSMarcoTRRetrieval & \textbf{15.83} & 8.37 & 7.52 & 7.11 \\
NFCorpusTR & \textbf{1.32} & 0.59 & 0.51 & 0.24 \\
QuoraRetrievalTR & \textbf{63.84} & 52.20 & 49.02 & 47.46 \\
SciFactTR & \textbf{23.97} & 21.19 & 19.86 & 17.34 \\
SquadTRRetrieval & \textbf{18.74} & 10.69 & 9.94 & 8.78 \\
TQuadRetrieval & \textbf{46.92} & 35.43 & 33.81 & 34.96 \\
TurkishAbstractCorpusClustering & \textbf{47.83} & 43.63 & 44.24 & 41.20 \\
XQuADRetrieval & \textbf{42.27} & 28.55 & 28.69 & 26.19 \\
\multicolumn{5}{l}{\textit{STS}} \\
STSbTR & \textbf{50.04} & 45.73 & 42.14 & 42.54 \\
\bottomrule
\end{tabular}
}
\end{table}

As shown in Table \ref{tab:mteb_detailed}, the TurkishTokenizer-based model achieved an average score of \textbf{39.57\%} across 26 TR-MTEB tasks, surpassing Mursit (35.92\%), CosmosGPT2 (35.65\%), and Tabi (34.92\%). Analyzing performance by category reveals distinct trade-offs. TurkishTokenizer demonstrates substantial advantages in \textit{STS} and \textit{Retrieval} tasks (e.g., TQuadRetrieval: 46.92\% vs 34.96\% for Tabi), which aligns with our hypothesis that morphology-aware segmentation improves the semantic quality of embeddings for similarity and search. At the same time, the Tabi baseline remains competitive or superior in specific \textit{Classification} tasks (e.g., Turkish75NewsClassification: 80.00\% vs 78.00\% for TurkishTokenizer), suggesting that different tokenization priors can favor different downstream regimes even under matched architecture and training protocol.

While TR-MTEB provides broad downstream coverage, it does not isolate controlled grammatical manipulations. We therefore complement it with TurBLiMP, which probes specific linguistic phenomena via minimal pairs.

TurBLiMP provides Turkish minimal pairs designed to probe specific linguistic phenomena (e.g., agreement, scrambling, nominalization) \cite{basar_turblimp_2025}. Since our models are sentence embedding encoders (rather than generative language models), we evaluate a centroid-based acceptability proxy: for each phenomenon file, we compute the centroid of the grammatical sentence embeddings, score each sentence by cosine similarity to this centroid, and count a minimal pair as correct if the grammatical sentence receives a higher score than its ungrammatical counterpart. We report the resulting pairwise accuracy per phenomenon (in \%) in Table~\ref{tab:turblimp_detailed}. Following the visualization convention used in the TurBLiMP paper, cell background colors indicate relative performance, ranging from lower (red) to higher (green).

\begin{table}[H]
\centering
\caption{TurBLiMP centroid-based pairwise accuracy by linguistic phenomenon.}
\label{tab:turblimp_detailed}
\resizebox{\linewidth}{!}{
\input{turblimp_results_tables/results_table_centroid_colored}
}
\end{table}

Across many categories, TurkishTokenizer improves the separation of grammatical vs.\ ungrammatical minimal pairs under this proxy. A complementary evaluation of true grammatical sensitivity would require scoring minimal pairs with language model likelihood or a supervised acceptability classifier, which we leave for future work.

\FloatBarrier

%% file: turblimp_results_tables/results_table_centroid_colored.tex
\begin{tabular}{lrrrr}
\toprule
 & TurkishTokenizer & Mursit & CosmosGPT2 & Tabi \\
\midrule
Anaphor Agreement & \cellcolor[RGB]{243,202,95} \textbf{52.3} & \cellcolor[RGB]{255,194,97} 48.6 & \cellcolor[RGB]{255,198,99} 49.6 & \cellcolor[RGB]{255,198,99} 49.6 \\
Argument Str. Tran. & \cellcolor[RGB]{255,189,94} 47.4 & \cellcolor[RGB]{231,205,90} 54.7 & \cellcolor[RGB]{219,207,86} \textbf{57.0} & \cellcolor[RGB]{222,206,87} 56.3 \\
Argument Str. Ditr. & \cellcolor[RGB]{249,201,98} 51.0 & \cellcolor[RGB]{255,145,72} 36.4 & \cellcolor[RGB]{207,210,81} \textbf{59.3} & \cellcolor[RGB]{247,201,97} 51.4 \\
Binding & \cellcolor[RGB]{67,240,26} \textbf{86.7} & \cellcolor[RGB]{149,222,58} 70.7 & \cellcolor[RGB]{150,222,59} 70.5 & \cellcolor[RGB]{185,214,72} 63.6 \\
Determiners & \cellcolor[RGB]{1,254,0} \textbf{99.7} & \cellcolor[RGB]{18,250,7} 96.3 & \cellcolor[RGB]{32,247,12} 93.6 & \cellcolor[RGB]{212,209,83} 58.4 \\
Ellipsis & \cellcolor[RGB]{189,214,74} \textbf{62.9} & \cellcolor[RGB]{255,197,98} 49.3 & \cellcolor[RGB]{255,189,94} 47.3 & \cellcolor[RGB]{227,205,89} 55.4 \\
Irregular Forms & \cellcolor[RGB]{230,205,90} \textbf{54.9} & \cellcolor[RGB]{255,130,65} 32.7 & \cellcolor[RGB]{255,133,66} 33.3 & \cellcolor[RGB]{255,183,91} 45.9 \\
Island Effects & \cellcolor[RGB]{35,247,13} \textbf{93.1} & \cellcolor[RGB]{255,199,99} 49.9 & \cellcolor[RGB]{255,142,71} 35.7 & \cellcolor[RGB]{251,200,98} 50.7 \\
Nominalization & \cellcolor[RGB]{255,181,90} 45.4 & \cellcolor[RGB]{248,201,97} 51.2 & \cellcolor[RGB]{246,201,96} 51.7 & \cellcolor[RGB]{244,202,95} \textbf{52.1} \\
NPI Licensing & \cellcolor[RGB]{143,224,56} \textbf{71.9} & \cellcolor[RGB]{255,187,93} 46.8 & \cellcolor[RGB]{255,194,97} 48.7 & \cellcolor[RGB]{255,193,96} 48.3 \\
Passives & \cellcolor[RGB]{255,167,83} 41.9 & \cellcolor[RGB]{232,204,91} 54.4 & \cellcolor[RGB]{255,172,86} 43.1 & \cellcolor[RGB]{196,212,77} \textbf{61.5} \\
Quantifiers & \cellcolor[RGB]{255,102,51} 25.5 & \cellcolor[RGB]{255,179,89} \textbf{44.9} & \cellcolor[RGB]{255,84,42} 21.1 & \cellcolor[RGB]{255,79,39} 19.8 \\
Relative Clauses & \cellcolor[RGB]{255,193,96} 48.4 & \cellcolor[RGB]{255,194,97} 48.7 & \cellcolor[RGB]{247,201,97} \textbf{51.5} & \cellcolor[RGB]{255,197,98} 49.4 \\
Scrambling & \cellcolor[RGB]{208,210,81} \textbf{59.1} & \cellcolor[RGB]{221,207,87} 56.5 & \cellcolor[RGB]{227,205,89} 55.3 & \cellcolor[RGB]{226,206,88} 55.6 \\
Subject Agreement & \cellcolor[RGB]{161,220,63} \textbf{68.4} & \cellcolor[RGB]{219,207,86} 56.9 & \cellcolor[RGB]{228,205,89} 55.1 & \cellcolor[RGB]{199,211,78} 60.8 \\
Suspended Affixation & \cellcolor[RGB]{146,223,57} \textbf{71.2} & \cellcolor[RGB]{255,181,90} 45.4 & \cellcolor[RGB]{255,161,80} 40.4 & \cellcolor[RGB]{255,189,94} 47.3 \\
Model Average & \cellcolor[RGB]{197,212,77} \textbf{61.2} & \cellcolor[RGB]{241,202,94} 52.7 & \cellcolor[RGB]{250,200,98} 50.8 & \cellcolor[RGB]{246,201,96} 51.6 \\
\bottomrule
\end{tabular}

%% file: chapters/future_work.tex
\section{Future Work}
\label{sec:future_work}

This study highlights the importance of linguistic integrity and computational efficiency in tokenization, presenting a framework to guide the development of tokenizers optimized for morphologically rich and low-resource languages. Despite these promising results, much work remains to unlock the full potential of tokenizers. Future improvements will focus on incorporating advanced morphological analysis steps, which will further enhance their capability to capture the rich grammatical and semantic structures of Turkish. These steps may include integrating more sophisticated linguistic rules, handling rare morphemes, and accounting for contextual variations that impact tokenization in complex languages. Such enhancements will not only improve linguistic fidelity but also expand the scope of the tokenizers for diverse NLP applications.

Future work will also explore adapting the tokenizer to additional languages. Extending the approach beyond Turkish requires constructing language-specific lexical resources (e.g., root and affix inventories) and corresponding decoding and normalization rules, and validating the resulting tokenizers on language-appropriate benchmarks.

Although still in the early stages of development, this tokenizer provides a strong foundation for further innovation. Its initial performance gives hope that, with targeted improvements, it can evolve into a robust, versatile tool for tokenizing morphologically rich languages. By implementing these additional steps and conducting further evaluations across languages and tasks, this research aims to establish a new standard for linguistically informed tokenization, ultimately advancing the quality and efficiency of language models in a wide array of applications.

%% file: chapters/conclusion.tex
\section{Conclusion}
\label{sec:conclusion}

We presented a linguistically informed, morphology-first hybrid tokenizer designed for Turkish and similar agglutinative languages. The tokenizer combines curated root and affix lexicons with phonological normalization (mapping surface allomorphs to shared identifiers) and a controlled subword fallback for coverage. This design aims to produce token sequences that more closely align with morpheme boundaries while remaining practical for large-scale NLP pipelines.

On TR-MMLU, the proposed tokenizer achieves 90.29\% TR~\% and 85.80\% Pure~\%, indicating substantially stronger morpheme-level alignment than several general-purpose tokenizers. We additionally report downstream sentence embedding evaluation on STS and TR-MTEB using \textbf{randomly initialized} models to isolate tokenizer effects from pretrained knowledge. The TurkishTokenizer-based model reaches \textbf{51.44\%} Pearson correlation on STSb-TR, compared to 43.01\% for the Tabi baseline-a gain of \textbf{+8.43 percentage points}. On TR-MTEB, TurkishTokenizer achieves 39.57\% overall average compared to 34.92\% for Tabi (+4.65 points). These gaps demonstrate that morphology-first tokenization provides a stronger inductive bias for learning Turkish semantic representations from scratch, and the same tokenizer also yields the strongest average accuracy on a centroid-based TurBLiMP minimal-pairs proxy.

We emphasize that empirical claims in this paper are Turkish-focused. We outline concrete next steps-improved morphophonological handling, better capitalization edge cases, and standardized efficiency measurements-in Section~\ref{sec:future_work}.